\documentclass{article}
\DeclareUnicodeCharacter{2606}{\ensuremath{\star}}

\usepackage[12pt]{extsizes}
\usepackage[utf8]{inputenc}
\usepackage{amsmath}
\usepackage{amssymb}
\usepackage{amsthm}
\usepackage{amsfonts}
\usepackage[numbers,sort&compress]{natbib}
\usepackage{graphicx}
\usepackage{lipsum}
\usepackage{enumitem}
\usepackage{minted}
\usepackage[font=footnotesize,labelfont=bf]{caption}
\usepackage{mdframed}
\usepackage{authblk}
\usepackage{longtable}
\usepackage{subfigure}
\usepackage[subfigure]{tocloft}
\usepackage[dvipsnames,table]{xcolor}
\usepackage[hidelinks]{hyperref}
\usepackage{mathtools}
\usepackage{relsize}
\usepackage{multirow}
\usepackage{booktabs}
\usepackage{tabularx}
\usepackage{bbding}
\usepackage{listings}
\usepackage{xspace}
\usepackage[ruled,vlined]{algorithm2e}
\usepackage{multirow}
\usepackage[p,osf]{cochineal}
\usepackage[scale=.95,type1]{cabin}
\usepackage[cochineal,bigdelims,cmintegrals,vvarbb]{newtxmath}
\usepackage[zerostyle=c,scaled=.94]{newtxtt}
\usepackage[cal=boondoxo]{mathalfa}
\usepackage{microtype}
\usepackage{multirow}
\usepackage{adjustbox}
\usepackage{etoolbox}
\usepackage{lmodern}
\usepackage{csquotes}

\usepackage{caption}
\usepackage{geometry}
\definecolor{myurlcolor}{HTML}{123463}
\definecolor{dc_color}{RGB}{230, 245, 244}
\definecolor{ds_color}{RGB}{195, 230, 227}
\definecolor{ms_color}{RGB}{150, 214, 209}
\hypersetup{
    colorlinks=true,
    linkcolor=black,
    filecolor=black,      
    urlcolor=black,
    citecolor=black
}

\apptocmd{\thebibliography}{\raggedright}{}{}

\makeatletter
\patchcmd{\@maketitle}{\LARGE \@title}{\fontsize{30}{19.2}\selectfont\@title}{}{}
\makeatother
\newcommand{\hide}[1]{}

\usepackage{cleveref}
\Crefname{section}{Sec.}{Secs.}
\Crefname{equation}{Eq.}{Eqs.}
\Crefname{figure}{Fig.}{Figs.}
\Crefname{tabular}{Tab.}{Tabs.}

\def\eqref#1{equation~\ref{#1}}

\DeclareMathAlphabet{\mathsfit}{\encodingdefault}{\sfdefault}{m}{sl}
\SetMathAlphabet{\mathsfit}{bold}{\encodingdefault}{\sfdefault}{bx}{n}

\usepackage{placeins}      
\usepackage{dblfloatfix}   
\usepackage[font=small,labelfont=bf]{caption} 

\usepackage{longtable,booktabs,array,ragged2e}
\usepackage[most]{tcolorbox}

\newcolumntype{L}[1]{>{\RaggedRight\arraybackslash}p{#1}}
\newcolumntype{C}[1]{>{\Centering\arraybackslash}p{#1}}

\providecommand{\nbeTableStyleOn}{}
\providecommand{\nbeTableStyleOff}{}

\newenvironment{RowLinesLT}[1]{%
  \begin{longtable}{#1}%
}{%
  \end{longtable}%
}

\newcounter{boxctr}
\tcbset{
  colback=white, boxrule=0.6pt, sharp corners,
  left=6pt, right=6pt, top=8pt, bottom=6pt,
  before skip=10pt, after skip=10pt
}
\newenvironment{nbeBox}[1]{%
  \refstepcounter{boxctr}%
  \begin{tcolorbox}[breakable,enhanced]%
  \noindent{\bfseries Box~\theboxctr\ |~#1}\\[2pt]%
  \noindent\makebox[\linewidth]{\rule{\linewidth}{0.4pt}}\\[-2pt]%
  \footnotesize%
}{%
  \end{tcolorbox}%
}

\makeatletter
\renewenvironment{nbeBox}[1]{%
  \refstepcounter{boxctr}%
  \par\bigskip
  \noindent{\bfseries Box~\theboxctr\ |~#1}\par
  \vspace{4pt}%
  \noindent\makebox[\linewidth]{\rule{\linewidth}{0.3pt}}%
  \vspace{6pt}%
  \nbeTableStyleOn
  \footnotesize
}{%
  \nbeTableStyleOff
  \par\vspace{6pt}%
  \noindent\makebox[\linewidth]{\rule{\linewidth}{0.3pt}}%
  \bigskip
}
\makeatother

\title{\LARGE\textbf{Foundation Model in Biomedicine}}
\author[1,$\ast$]{Xiangrui Liu} 
\author[2,$\ast$]{Yuanyuan Zhang} 
\author[3,$\ast$]{Qianyu Shang} 
\author[4]{Yingzhou Lu} 
\author[5]{Changchang Yin} 
\author[6,7]{Xiaoling Hu} 
\author[1]{Xiaoou Liu} 
\author[8]{Lulu Chen} 
\author[9]{Alexander Rodr{\'\i}guez} 
\author[1]{Yezhou Yang} 
\author[5]{Ping Zhang} 
\author[10]{Jintai Chen} 
\author[3]{Shan Du} 
\author[11,$\#$]{Huaxiu Yao} 
\author[12,$\#$]{Sheng Wang} 
\author[13,$\#$]{Tianfan Fu} 
\author[12,$\#$]{Xiao Wang} 

\affil[1]{ Arizona State University, Tempe, AZ, USA}
\affil[2]{ Purdue University, West Lafayette, IN, USA}
\affil[3]{ University of British Columbia, Kelowna, BC, Canada}
\affil[4]{ Stanford University, Stanford, CA, USA}
\affil[5]{ The Ohio State University, Columbus, OH, USA}
\affil[6]{ Massachusetts General Hospital, Boston, MA, USA}
\affil[7]{ Harvard Medical School, Boston, MA, USA}
\affil[8]{ Virginia Polytechnic Institute, Blacksburg, VA, USA}
\affil[9]{ University of Michigan, Ann Arbor, MI, USA}
\affil[10]{ HKUST Guangzhou, Guangzhou, Guangdong, China}
\affil[11]{ University of North Carolina at Chapel Hill, Chapel Hill, NC, USA}
\affil[12]{ University of Washington, Seattle, WA, USA}
\affil[13]{ Nanjing University, Nanjing, Jiangsu, China}

\affil[$\#$]{\em{Corresponding authors: \href{mailto:wang3702@uw.edu}{wang3702@uw.edu}, \href{mailto:futianfan@nju.edu.cn}{futianfan@nju.edu.cn}, 
\href{mailto:swang@cs.washington.edu}{swang@cs.washington.edu}, \href{mailto:huaxiu@cs.unc.edu}{huaxiu@cs.unc.edu} }}
\affil[$\ast$]{\em{These authors contributed equally to this work.}}

\date{}
\begin{document}
\maketitle
\begin{abstract}
Foundation models, first introduced in 2021, refer to large-scale pretrained models (e.g., large language models (LLMs) and vision-language models (VLMs)) that learn from extensive unlabeled datasets through unsupervised methods, enabling them to excel in diverse downstream tasks. These models, like GPT, can be adapted to various applications such as question answering and visual understanding, outperforming task-specific AI models and earning their name due to broad applicability across fields. 
The development of biomedical foundation models marks a significant milestone in the use of artificial intelligence (AI) to understand complex biological phenomena and advance medical research and practice. 
This survey explores the potential of foundation models in diverse domains within biomedical fields, including computational biology, drug discovery and development, clinical informatics, medical imaging, and public health. 
The purpose of this survey is to inspire ongoing research in the application of foundation models to health science.
\end{abstract} 

\section{Introduction}
\label{sec:intro}

The term ``foundation model'', first introduced in 2021 \cite{bommasani2021opportunities}, generally refers to large language models (LLMs) and vision language models (VLMs) that are pre-trained in large-scale datasets, usually through unsupervised methods, which equip them to handle diverse downstream tasks. 
By learning from vast amounts of unlabeled data, ``foundation models'' have developed strong capacities to map inputs into latent embedding space. 
Consequently, they can be seamlessly adapted to a wide range of tasks, consistently outperforming task-specific AI models \cite{awais2025foundation,zhou2024comprehensive}.
For example, GPT \cite{brown2020language}, pre-trained on massive language and visual data, achieves outstanding performance in numerous tasks such as question answering, information retrieval, and visual understanding.
Given their transformative potential and broad applicability across related fields, these models are commonly referred to as ``foundation models''.

The emergence and development of foundation models can be attributed to several key factors. 
\textbf{Massive unlabeled data}: Vast amounts of data are available, but supervised training is impractical due to prohibitive labeling costs~\cite{bommasani2021opportunities}.
\textbf{Increased AI model size}: The architectures of the AI model have evolved to become increasingly larger, but the limited availability of labeled data constrains their ability to fully exploit this enhanced capacity \cite{yuan2023power}.
\textbf{Scaling law of generalizability}: Through large-scale model training, researchers have found that model performance improves predictably with increases in model size, dataset size, and computational resources~\cite{kaplan2020scaling}.
\textbf{Cost-efficient for downstream tasks}: After pre-training, efficient fine-tuning with limited labeled data achieves superior performance compared to task-specific AI models.



The success of popular foundation models such as GPT and Claude in natural language and image processing makes it intuitive to apply and redesign them to healthcare. 
The application of foundation models in healthcare spans several sub-fields. First, the outstanding natural language processing capabilities of foundation models have the potential to advance computational biology. DNA, RNA and protein sequences can be seen as a form of natural language, and these models can learn the patterns in the sequences, enabling deeper insights into genomics. 
Second, drug discovery and development utilizes foundation models to accelerate target identification, optimize molecular design, and predict molecular interactions and properties, ultimately reducing the time and cost of developing new drugs \cite{mak2024artificial}. 
Third, in clinical informatics, foundation models can efficiently process millions, or even billions, of clinical and patient data points, whether structured or unstructured. 
They can extract patterns from patients’ symptoms to better assess conditions and enable personalized treatment plans. 
Fourth, medical imaging analysis can employ foundation models for tasks such as image segmentation, anomaly detection, and diagnostic predictions across modalities such as MRI and CT \cite{zhang2024challenges}, improving diagnostic accuracy and workflow efficiency. 
Public health benefits from foundation models in analyzing large datasets for disease surveillance, epidemiological modeling, and misinformation detection, contributing to more effective public health interventions. Therefore, the opportunities for biomedical foundation models to enhance the work of clinicians, researchers, and patients are steadily increasing. 



\begin{figure*}[!t]
    \centering
    \includegraphics[width=1\linewidth]{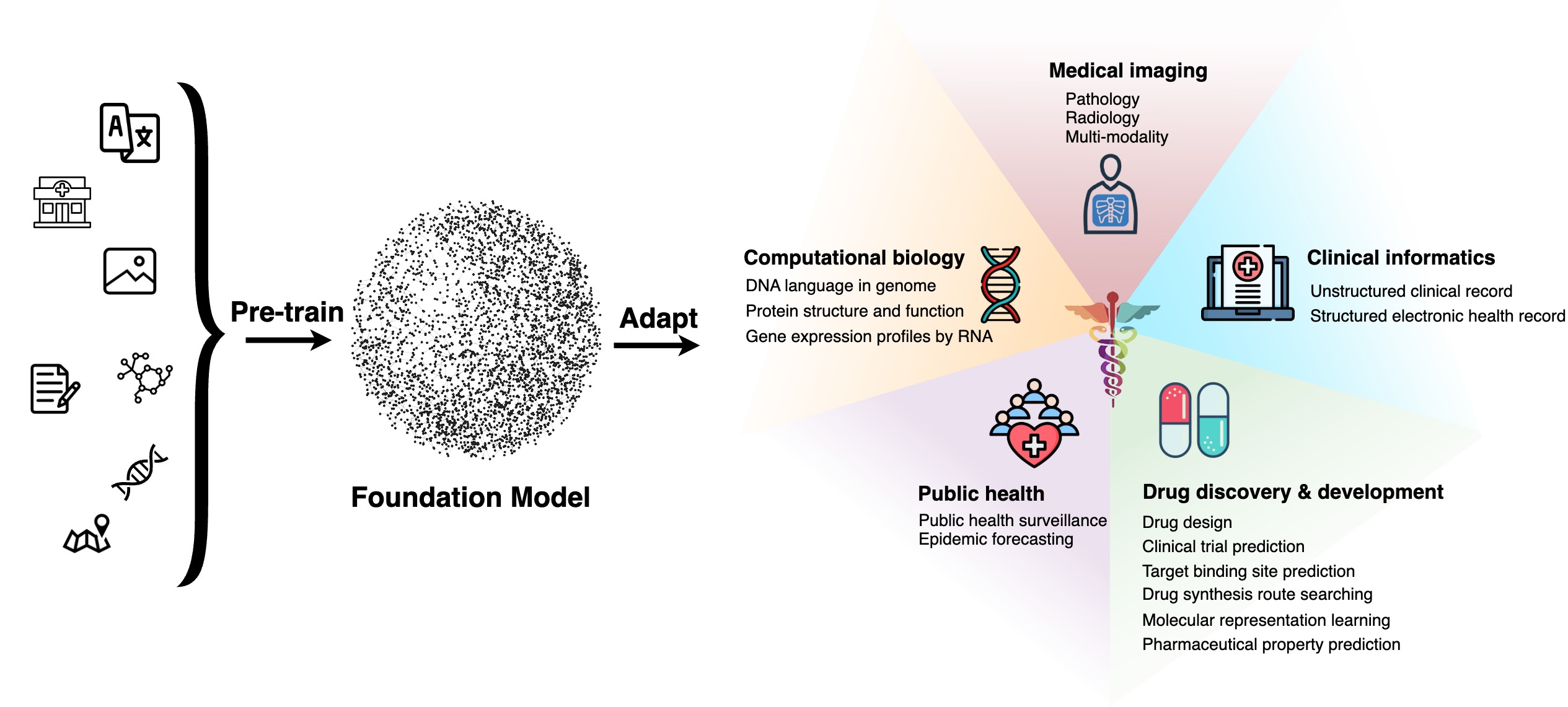}
    \caption{\textbf{Overview of the foundation models in different biomedical fields.}
    The foundation model is first pre-trained with massive unlabeled data in a self-supervised fashion. 
    Then, it can be easily adapted for various downstream applications, including computational biology, drug discovery, public health, medical imaging, and clinical informatics. }
    \label{fig:main}
\end{figure*}

This survey aims to review existing research on foundation models in biomedical areas, summarize their development progress, identify recent challenges of biomedical foundation models to inspire potential research directions and provide a foundation for researchers to advance their applications in health sciences. 
Specifically, we will discuss the foundation models in multiple biomedical fields, including computational biology, drug discovery and development, clinical informatics, medical imaging, and public health (\textbf{Figure~\ref{fig:main}}).


\begin{figure*}[!t]
    \centering
    \includegraphics[width=1\linewidth]{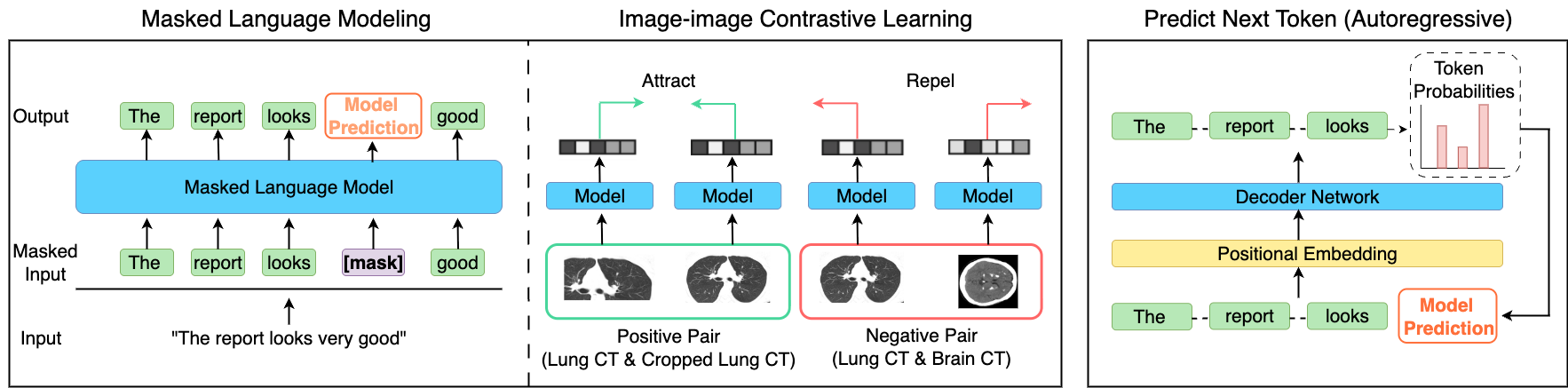}
    \caption{\textbf{Overview of foundation models training strategies including masked language modeling for token recovery, contrastive learning for aligning representations across image pairs, and next-token prediction for autoregressive sequence modeling.}
    }
    \label{fig:FM_train}
\end{figure*}

\section{Foundation Model}
\label{sec:foundation_model}
To better understand these foundation models, we categorize them into two primary types: discriminative models and generative models. Discriminative models focus on the decision boundary between classes and are more directly concerned with prediction or classification tasks. In contrast, generative models aim to understand and replicate the data distribution, allowing them to generate new data instances similar to the training data. 

\paragraph{Discriminative Foundation Model}

 The discriminative foundation model generally refers to the group of foundation models that are specifically trained to differentiate between inputs through unsupervised learning. The term ``discriminative'' in this context means that the model focuses on learning the differences between different classes or categories of data. Masked language modeling (MLM) is a well-known training strategy for discriminative foundation models. During MLM training, some tokens in the input text are randomly replaced with a special token, and the model learns to predict the original tokens. This approach forces the model to understand contextual relationships and encodes bidirectional information, making it effective for learning language representations. A classic example is BERT \cite{devlin2019bert}, which is the prototype backbone for many biomedical foundation model variants that have textual inputs. Another popular strategy is contrastive learning which trains a model to embed “positive” pairs (e.g., matching text–image pairs or two augmented views of the same image) closer together in the embedding space while segregating “negative” pairs. A successful case of contrastive learning based foundation model would be the CLIP \cite{radford2021learning} which built the foundation for many modern VLMs. 
 
\paragraph{Generative Foundation Model}
While PLMs typically possess a task-agnostic architecture, they often require fine-tuning to adapt to downstream tasks effectively. To address this issue, researchers propose Autoregressive (AR) modelling for better few-shot and even zero-shot performance. These AR language models are trained by predicting the next word in a sequence based on the preceding words---``predict the next token''. As one of the milestones, GPT~\cite{Brown2020LanguageMA} utilizes the decoder-only transformer architecture to predict the next words, exhibiting excellent performance.

\begin{figure}[!t]
  \centering
  \includegraphics[width=\textwidth,height=1.3\textheight,keepaspectratio]{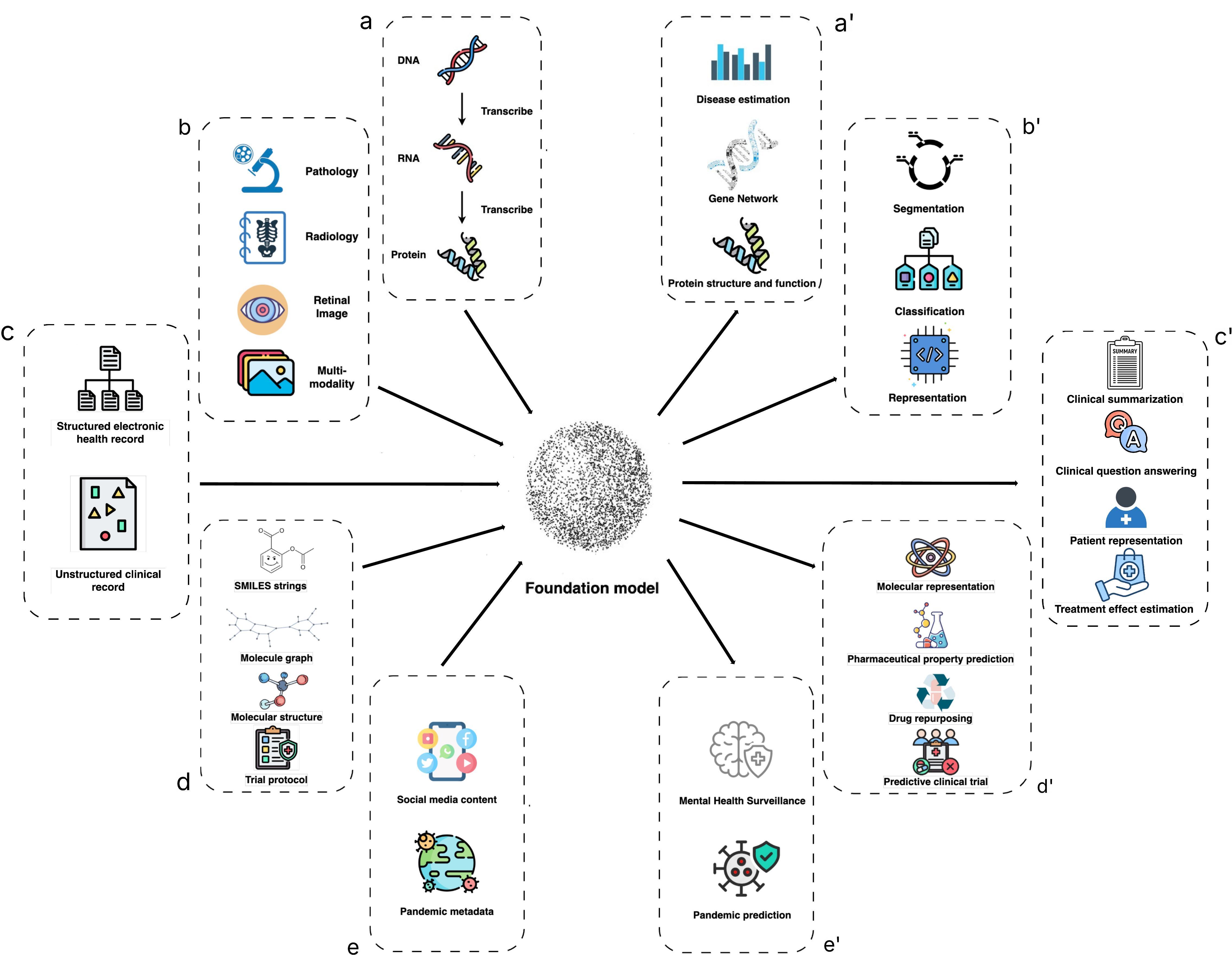}
  \caption{\textbf{Applications of foundation models across biomedical domains.}
  \textbf{(a)} Computational biology: foundation models learn from DNA, RNA, and protein sequences for disease estimation, gene networks, and protein structure prediction.
  \textbf{(b)} Medical imaging: pathology, radiology, and retinal images enable segmentation, classification, and multimodal representation learning.
  \textbf{(c)} Clinical informatics: structured and unstructured health records support summarization, question answering, patient representation, and treatment effect estimation.
  \textbf{(d)} Drug discovery and development: molecular representations drive property prediction, drug repurposing, and predictive clinical trials.
  \textbf{(e)} Public health: multimodal signals from social media and epidemic metadata enable mental health surveillance and pandemic forecasting.  
  }
  \label{fig:biomedical_applications}
\end{figure}

To provide a unifying overview of how foundation models are being applied across biomedical domains, we present an integrated schematic (Figure~\ref{fig:biomedical_applications}) that highlights five major application areas: computational biology, drug discovery, clinical informatics, medical imaging, and public health.

\FloatBarrier

\section{Computational Biology}
\label{sec:computational_biology}

Computational biology plays a central role in translating the vast and complex data from modern sequencing and molecular profiling technologies into actionable biological insights. Fundamental tasks in this domain, such as understanding genome structure, interpreting RNA expression, and predicting protein structures, are often hindered by data heterogeneity, sparse annotations, and the intricate nature of biological systems. Traditional, heuristic-based approaches, while valuable, frequently lack the generalizability required to adapt to new datasets or contexts.

Foundation models offer a transformative alternative. By pretraining on extensive biological sequences, structures, or multimodal data, these models acquire generalized representations that encapsulate fundamental biological principles. This enables them to be efficiently fine-tuned for a wide range of downstream applications, from analyzing different species to tackling diverse tasks with minimal domain-specific modifications. As illustrated in \textbf{Figure~\ref{fig:biomedical_applications}a}, foundation models in computational biology span multiple levels of molecular information, from genome to transcriptome to proteome.

\paragraph{Genome Information}
Genome analysis is a cornerstone of modern biology, providing the foundation for understanding the genetic underpinnings of health and disease. Traditionally, this process has relied on a series of specialized bioinformatics pipelines which often use domain-specific heuristics and hand-crafted features, limiting their generalizability to diverse genomic data or species. For instance, methods for predicting non-coding regulatory elements such as enhancers are frequently tailored to specific cell types and may fail to generalize across different biological contexts.

Foundation models offer a transformative alternative by treating DNA as a language. Through pretraining on raw genomic sequences, these models learn latent representations that capture the complex statistical and functional relationships between nucleotides, including long-range dependencies. Transformer-based models are particularly well-suited for this task due to their ability to process long input contexts. Models in this space vary in their architecture and scale. For example, Nucleotide Transformer~\cite{dalla2023nucleotide} and DNABERT-2~\cite{liu2024dnabert2} apply MLM over millions of sequences to support genome-wide annotation and variant effect prediction. Others, such as Enformer~\cite{avsec2021effective}, use hybrid convolution and attention layers to model large genomic windows (up to 200kb) for chromatin and expression prediction. Meanwhile, HyenaDNA~\cite{poli2024hyenadna} improves scalability by replacing traditional attention with efficient long-range operators. These models collectively represent a significant shift toward data-driven genome interpretation that scales with available sequencing data. Representative genome-level foundation models and associated datasets are summarized in Box~\ref{box:genome}.

\paragraph{Gene Expression Profiles by RNA}
Gene expression profiling allows researchers to monitor cellular states and responses across various conditions. While technologies like RNA-seq enable genome-wide quantification of expression levels, analyzing these data remains challenging due to noise, batch effects, and tissue variability. Traditional statistical and dimensionality reduction techniques often lack transferability across datasets.

Foundation models provide a new paradigm for modeling gene expression data. By pretraining on large single-cell or bulk RNA sequence datasets, they learn expression-level features that are generalized across different tissues or experimental setups. These models, often built with transformer architectures, are typically trained with objectives like cell-type classification, denoising, or expression imputation. Notable efforts in this area have demonstrated significant progress. scGPT~\cite{zhang2023scgpt} is trained on single-cell profiles and can perform downstream tasks such as clustering and perturbation prediction. xTrimoGene~\cite{chen2023xtrimogene} focuses on scalability and transfer learning across datasets, while Geneformer~\cite{Geneformer} and GenePT~\cite{gupta2024genept} build generalized embeddings from large transcriptomic corpora. These models improve downstream applications like cell type annotation and disease state prediction by providing unified representations across a multitude of data sources. Key models and benchmarks for RNA-based representation learning are outlined in Box~\ref{box:rna}.

\paragraph{Protein Structure and Protein Design}
Proteins are the workhorses of biology, and understanding their three-dimensional structure is key to decoding their function. However, solving protein structures experimentally is both costly and time-consuming. While early predictive models offered a promising alternative, they often lacked the accuracy required to replace lab-based methods.

Foundation models have revolutionized this field by treating protein sequences as a language and implicitly learning folding rules from vast corpora of sequences and structures. The attention mechanism in transformers allows these models to capture the complex spatial dependencies between residues, a critical factor for accurate structure prediction. The most iconic example is AlphaFold2~\cite{jumper2021highly}, which achieved unprecedented accuracy in single-chain structure prediction. Subsequent models like ESMFold~\cite{lin2022language} and OmegaFold~\cite{wu2022high} streamlined the inference process by removing the need for multiple sequence alignments. On the design front, ProGen2~\cite{nijkamp2022progen} and Protein Generator~\cite{shin2024protein} use generative transformers to create novel sequences with desired properties. This paradigm has been further advanced by AlphaFold3~\cite{alley2024alphafold3}, which extends prediction capabilities to include multimers and protein-ligand interactions. Together, these models are shifting protein science toward a data-centric, generative approach with significant implications for therapeutics and synthetic biology. A concise summary of protein structure and design foundation models is provided in Box~\ref{box:protein}.

\section{Medical Imaging}
\label{sec:med_image}

Medical imaging plays a pivotal role in modern healthcare, encompassing a wide array of technologies meticulously designed to visualize distinct aspects of the human body. These modalities are instrumental in diagnosing and monitoring various medical conditions, as well as evaluating treatment efficacy. Ranging from identifying potential injuries or diseases to gauging the progression of a condition, these technologies provide diverse clinical insights~\cite{suetens2017fundamentals, beutel2000handbook}. However, a primary challenge in developing advanced medical imaging models is the scarcity of suitable data. The medical visual examination involves a variety of images, such as radiology and pathology, yet much of this data is private. This makes it difficult to collect sufficiently large and diverse datasets to train general foundation models, a requirement for their effective generalization across different modalities and tasks.
As summarized in \textbf{Figure~\ref{fig:biomedical_applications}b}, foundation models for medical imaging span radiology, pathology, and retinal imaging, and support core tasks such as segmentation, classification, and representation learning across modalities.

\paragraph{Image Classification}
Image classification is a fundamental task in medical image analysis, used to identify the presence of diseases, grade their severity, or categorize image types. While traditional Convolutional Neural Networks (CNNs) have shown promise, they often struggle with generalization across different institutions or image modalities, particularly when labeled data is scarce. This lack of transferability limits their utility in real-world clinical settings.

Foundation models address these limitations by leveraging a broader understanding of visual features gained from pretraining on vast, often cross-domain or multimodal, datasets. By learning from both images and associated text, these models develop robust representations that can be effectively transferred to medical tasks, even with limited annotated data. This approach also enables powerful zero-shot or few-shot learning, which is particularly valuable in scenarios where acquiring labeled data is costly or time-consuming. In this domain, models such as CheXzero~\cite{tiu2022expert} have demonstrated that zero-shot classification using a CLIP-based architecture can match the performance of expert radiologists in detecting chest diseases. RETFound~\cite{zhou2023foundation} is another notable example that was pretrained on millions of retinal images and shows strong performance across various eye disease classification tasks. Furthermore, models like MedCLIP~\cite{zhang2023biomedclip} utilize contrastive learning between radiology images and their corresponding report texts to support general classification and retrieval. These works collectively highlight how pretraining can make classification tasks more flexible and data-efficient. Representative foundation models for medical image classification are summarized in Box~\ref{box:retinal_image}.

\paragraph{Image Segmentation}
Image segmentation involves identifying and outlining specific regions in a medical image, such as tumors, organs, or lesions. This task is crucial for numerous clinical applications, including measuring tumor size, surgical planning, and tracking disease progression. However, acquiring high-quality segmentation data is challenging due to the time-consuming nature of manual annotation, which requires specialized domain expertise and high-resolution labeling.

Foundation models are well-suited to this task because they can learn generalized visual patterns from large unlabeled datasets and then be adapted for segmentation with minimal supervision. Some of these models also support prompt-based or instruction-based tuning, allowing users to interactively segment specific anatomical structures, such as "the left lung" or "a tumor area," without extensive retraining. Their inherent generality makes them adaptable across diverse tasks and medical domains. Examples include MedSAM~\cite{ma2024segment}, a medical adaptation of the Segment Anything Model, which performs prompt-based segmentation across different modalities. Jaume et al.~\cite{jaume2025multistain} further demonstrate the utility of multistain pretraining to enhance the generalization of pathology segmentation across different tissue types. These models significantly reduce the need for heavy supervision and enable more interactive and scalable segmentation workflows. Core pathology-oriented segmentation foundation models and resources are summarized in Box~\ref{box:pathology}.

\paragraph{Multi-task Learning}
In clinical practice, models are often required to perform more than a single task. Integrated systems that combine tasks like disease detection, localization, and explanation are highly valued. Multi-task learning offers a solution by training a single model to perform multiple functions simultaneously, which often leads to improved performance and simplifies deployment.

Foundation models are a natural fit for this paradigm due to their strong general representations. Since they are pretrained on a variety of tasks or data sources, they can support classification, segmentation, detection, and even captioning within a single unified architecture. Many of these models use a shared encoder with lightweight, task-specific heads, which enhances computational efficiency and simplifies the training process. UNI~\cite{chen2024towards}and MaCO~\cite{huang2024enhancing} are good examples of models that perform multiple tasks, such as segmentation and report generation, using a shared visual backbone. ELIXR~\cite{xu2023elixr}  addresses temporal radiology tasks by modeling sequences of imaging studies. MedCLIP~\cite{wang2022medclip} is also frequently used as a foundational backbone for building multi-task systems. These models demonstrate the growing feasibility of developing all-in-one solutions for comprehensive medical image understanding. Radiology-focused multi-task foundation models that unify detection, segmentation, classification, and reporting are summarized in Box~\ref{box:Radiology}.

\paragraph{Cross-modal Generation}
Some medical imaging tasks extend beyond mere recognition, requiring the model to reason about or explain the visual content. Examples include writing radiology reports, answering questions based on an image, or describing abnormalities on a scan. These tasks are particularly challenging as they demand a deep understanding of both visual and textual information and the ability to link them in a meaningful way.

Foundation models, especially those trained on image-text pairs, are highly effective in bridging this vision-language gap. They are pretrained to match images with descriptions, generate captions, or follow instructions grounded in visual content. Recent developments in instruction tuning and multimodal prompting have enabled these models to answer complex medical questions based on images and even simulate doctor-patient interactions. Med-Flamingo~\cite{alayrac2022flamingo} and LLaVA-Med \cite{li2024llava} are specifically designed for image-grounded text generation and question answering. BiomedGPT~\cite{zhang2023biomedgpt} can perform multiple reasoning tasks, including captioning and retrieval. In the field of pathology, PathChat \cite{lu2023foundational} supports dialogue-based interaction with whole-slide images. These models are making medical AI systems more interpretable and interactive, which holds significant potential for improving patient communication and supporting clinical training. Vision–language foundation models for image-grounded reasoning and generation are summarized in Box~\ref{box:medical_imaging}.

\section{Clinical Informatics}
\label{sec:clinical_health}

Clinical informatics encompasses both structured and unstructured health data, including patient medical histories, treatments, and care pathways. These records are crucial for supporting clinical decisions, enabling patient monitoring, and ultimately improving health outcomes. Foundational tasks in this domain are broadly grouped into two categories: text-centric tasks, such as summarization and question answering, and record-centric tasks, which include patient representation and treatment effect estimation. As illustrated in \textbf{Figure~\ref{fig:biomedical_applications}c}, foundation models in clinical informatics integrate both structured and unstructured data to support these diverse applications.

\paragraph{Clinical Summarization}
Clinical summarization generates concise overviews from lengthy medical records, which helps reduce physician workload and improve continuity of care. Unlike general summarization, this task must address challenges such as irregular documentation, domain-specific language, and maintaining factual consistency under limited annotations.

Foundation models support this task by leveraging biomedical pretraining to understand medical semantics and temporal context. Their capacity to integrate structured and unstructured data allows for comprehensive summary generation across different patient data types. Recent methods include extractive models such as BioBERTSum~\cite{du2020biomedical} and MRC-Sum~\cite{li2023mrc}, which identify salient sentences. Adapter-based approaches like KeBioSum~\cite{xie2022pre} incorporate domain knowledge into transformers, while COVIDSum~\cite{cai2022covidsum} applies graph-aware summarization for pandemic data. Radiology-LLaMA2~\cite{liu2023radiology} enables multimodal summarization using visual-textual alignment. Representative foundation models for clinical summarization are summarized in Box~\ref{box:summarization}.

\paragraph{Clinical Question Answering (QA)}
Clinical QA systems provide answers to medical queries based on clinical notes, guidelines, or literature. These tasks demand an accurate understanding of medical language, the ability to infer relevant information, and contextual reasoning over long documents.

Foundation models excel in QA due to their capacity for long-range dependency modeling and domain adaptation. Trained on medical corpora, they can generate relevant responses and adapt to different types of queries, whether extractive, generative, or dialog-based. Early models relied on BERT-based architectures~\cite{yoon2019pre,chakraborty2020biomedbert,rawat2020entity}. DAPO~\cite{li2023dialogue} improved QA through dialogue-specific modules. More recent generative models include ClinicalGPT~\cite{wang2023clinicalgpt}, PMC-LLaMA~\cite{wu2024pmc}, and Med-PaLM2~\cite{singhal2023towards}, which have achieved high performance across multiple Q\&A benchmarks. Representative foundation models for clinical summarization and question answering are summarized in Box~\ref{box:clinicalqa}.

\paragraph{Patient Representation}
Encoding patient records into meaningful representations is essential for modeling disease trajectories and supporting clinical decisions. This involves translating medical events, such as diagnoses, lab tests, and procedures, into numerical embeddings that capture both content and temporal order. 

Foundation models can encode diverse clinical events while handling time gaps and irregular sampling. Pretraining on large-scale EHR data makes these models more robust in sparse or noisy scenarios. BEHRT~\cite{li2020behrt} first introduced masked modeling to diagnosis sequences. Based on that, CEHR-BERT~\cite{pang2021cehr} extends the work by adding time-sensitive tokens, and ExBEHRT~\cite{rupp2023exbehrt} combined multimodal inputs to improve the results. Whereas, EHRMamba~\cite{fallahpour2024ehrmamba} adopted state-space models for efficient sequence modeling. Hierarchical models like Hi-BEHRT~\cite{li2022hi} and graph-aware models like GT-BEHRT~\cite{poulain2024graph} capture cross-visit and inter-variable relationships. Claimsformer~\cite{gerrard2024claimsformer} focuses on billing data for broader patient profiling. Key models for patient representation learning and treatment effect estimation are summarized in Box~\ref{box:patient_representation}.

\paragraph{Treatment Effect Estimation}
Treatment effect estimation (TEE) predicts how an intervention would affect a specific patient by learning from observational health data. This task supports personalized medicine and clinical trial design. Traditional methods require strong assumptions and are often challenged by confounding bias. 

Foundation models offer an alternative by learning contextualized patient histories and modeling interactions between treatments and covariates. These models are better suited to encode long-term dependencies and adjust for complex confounding structures. TransTEE~\cite{zhang2022exploring} uses a transformer backbone for estimating heterogeneous treatment effects across multiple treatment arms. CURE~\cite{liu2022cure} extends this with time-aware embeddings and pretraining on large observational datasets. These models demonstrate how pretrained representations can enhance causal inference tasks in healthcare. Foundation models designed for treatment effect estimation are summarized in Box~\ref{box:patient_representation}.

\section{Drug Discovery and Development}
\label{sec:drug_discovery}

The development of new drugs is vital for improving global health outcomes. The drug development pipeline is broadly divided into two major stages: early-stage discovery and late-stage development. The discovery phase focuses on identifying promising molecules with desirable pharmaceutical properties such as absorption and safety, either through rational design or repurposing. Once identified, these candidate molecules progress to the development phase, where they are rigorously tested in preclinical studies and clinical trials. Successful candidates are then submitted for regulatory review and potential approval. Recent advances in artificial intelligence, particularly foundation models, are fundamentally reshaping this pipeline. In early discovery, tasks such as molecular representation learning, property prediction, and drug repurposing are being transformed by large-scale pretraining. Correspondingly, in late-stage development, foundation models show significant promise in predicting clinical trial outcomes and optimizing trial design. As illustrated in \textbf{Figure~\ref{fig:biomedical_applications}d}, these applications collectively demonstrate how foundation models can reshape the entire drug development process.

\paragraph{Molecular Representation Learning}
Learning generalizable molecular representations is a fundamental challenge in drug discovery. Conventional methods, such as molecular fingerprints or graph kernels, often rely on hand-crafted rules and lack the flexibility to capture complex chemical patterns. With the increasing availability of large-scale chemical and biomedical datasets, foundation models trained on molecular structures or related text have emerged as a more adaptable solution. These models are typically trained using self-supervised objectives, including masked prediction, molecule-text alignment, or captioning. Depending on the input, they operate on SMILES sequences, molecular graphs, or paired molecule–text corpora. By capturing both fine-grained chemical context and broader structural information, these models can be adapted for various downstream tasks, such as molecule generation, property prediction, and reaction modeling.

Box~\ref{box:molecular} highlights key examples, which can be broadly categorized by their input modalities. Some models, such as MolFM~\cite{luo2023molfm}, use only molecular structures and employ multi-task training on graph input. MolKD~\cite{zeng2023molkd} builds on this by incorporating reaction yield information using knowledge distillation. Other approaches integrate chemical text with molecular input. BioT5~\cite{pei2023biot5}, CLAMP~\cite{seidl2023clamp}, and InstructMol~\cite{cao2023instructmol} follow a sequence-to-sequence framework, while BioT5+ extends this by incorporating SELFIES~\cite{selfies} (another sequence-based molecular descriptor, like SMILES) and chemical descriptions. Multimodal models, including MV-Mol~\cite{li2024mv} and UniMoT~\cite{zhang2024unimot}, combine chemical graphs with biomedical text and knowledge graphs. MoleculeSTM\cite{liu2023multi} adopts an instruction-tuning strategy to support a wide range of molecule-level tasks, signaling a broader shift toward general-purpose molecular language models.

\paragraph{Pharmaceutical Property Prediction}
Identifying chemical compounds with favorable pharmaceutical properties, collectively known as ADMET (Absorption, Distribution, Metabolism, Excretion, and Toxicity), is a crucial aspect of early-stage drug discovery. While traditional methods like rule-based filters or QSAR models are common, their performance often degrades on novel chemical scaffolds that fall outside their training data. Foundation models offer a more robust alternative by learning molecular representations from large chemical corpora. These pretrained models are capable of capturing both local atomic features and higher-level structural patterns, which significantly improves prediction performance, particularly in settings with limited labeled data.

Box~\ref{box:PPP} presents representative methods. BERT-style models, including ChemBERTa~\cite{ahmad2022chemberta}, SMILES-BERT~\cite{wang2019smiles}, and MFBERT~\cite{chen2022mfbert}, use masked language modeling to learn contextual representations from SMILES strings. Autoregressive approaches, such as DrugGPT~\cite{zhang2023druggpt}, treat SMILES as generative sequences, allowing for joint training of property prediction and molecule generation. SMILES-Mamba~\cite{xu2024smilesmamba} explores state-space architectures to reduce training overhead and improve efficiency. Other methods, such as MolE~\cite{su2022molecular}, ActFound\cite{feng2024bioactivity}, and GTFM\cite{mizera2024graph}, use graph-based transformers or meta-learning to operate directly on molecular graphs. These diverse approaches collectively reflect a growing architectural trend in pharmaceutical property prediction.

\paragraph{Drug Repurposing}
Drug repurposing, or repositioning, aims to identify new therapeutic uses for existing compounds. This approach offers significant advantages over developing new drugs from scratch, including lower costs, shorter timelines, and reduced risk. The primary challenge lies in uncovering meaningful associations between existing drugs and diseases that were not previously linked. Foundation models open new possibilities for this task by leveraging diverse inputs, ranging from biomedical literature and knowledge graphs to omics data, to uncover latent drug-disease connections. Some methods represent drug–target–disease relationships as graphs and use link prediction or node classification to prioritize candidates for repurposing.

Graph neural networks have become especially popular in this space, as they can effectively capture complex relational structures in biomedical graphs to predict novel indications for existing compounds~\cite{zhu2020knowledge, gao2022kg, bang2023biomedical, huang2024foundation}. 
As summarized in Box~\ref{box:repurposing}, most recent approaches, including Zhu et al.\cite{zhu2020knowledge}, KG-Predict~\cite{gao2022kg}, DREAMwalk~\cite{bang2023biomedical}, and TxGNN~\cite{huang2024foundation}, adopt GNN-based architectures to integrate diverse biomedical entities such as drugs, diseases, and genes. 
Notably, HGTDR~\cite{gharizadeh2024hgtdr} departs from the conventional GNN design by leveraging a Graph Transformer architecture, which enables more expressive modeling of heterogeneous interactions and improves the interpretability of drug–disease association predictions.

\paragraph{Predictive Clinical Trial}
Clinical trials are a critical, but often lengthy, expensive, and inefficient step in validating drug safety and efficacy. Bringing a drug to market can take over seven years and cost billions of dollars, with a low approval rate of around 15\%~\cite{martin2017much,huang2022artificial}. Failures frequently result from suboptimal trial design, such as poor patient stratification or inadequate endpoint definitions. Foundation models offer a way to improve trial design and predict trial outcomes. These models can integrate multiple data modalities, including trial protocols, patient data, and molecular profiles, to identify factors that correlate with successful trials. They can also simulate hypothetical scenarios to support clinical decision-making.

Recent works on clinical trial modeling can be grouped by architecture and task formulation. GNN-based models, such as HINT~\cite{fu2022hint} and HINT-UQ\cite{chen2024uncertainty}, encode molecular and disease relationships to predict trial outcomes. HINT-UQ extends this by modeling uncertainty for improved reliability. Transformer-based models, including inClinico~\cite{aliper2023prediction} and TrialDura~\cite{yue2024trialdura}, focus on outcome or duration prediction by learning temporal and protocol-aware representations from multimodal trial data. LIFTED~\cite{zheng2024multimodal} adopts a sparse mixture-of-experts design to identify cross-modal patterns and produce interpretable predictions. Another research direction explores language models like CTP-LLM~\cite{reinisch2024ctp} and ClinicalAgent~\cite{yue2024clinicalagent}, which leverage GPT and GPT-4 to reason over clinical documents and simulate agent-based trial planning. ClinicalAgent~\cite{yue2024clinicalagent} further incorporates multi-agent decision-making and instruction following for robust outcome estimation. TrialEnroll~\cite{yue2024trialenroll} uses a deep cross network architecture to predict patient enrollment success by modeling eligibility criteria and trial complexity. These diverse models reflect a growing emphasis on data integration and decision support in clinical development. Box~\ref{box:trial} summarizes the current approaches for trial outcome, duration, and enrollment forecasting.

\section{Public Health}
\label{sec:public_health}

Public health focuses on protecting community well-being through proactive interventions and population-level insights. The demand for timely and precise responses has increased following major infectious disease outbreaks such as COVID-19 and H1N1. Accurate forecasting and monitoring tools are essential to inform decision-makers during epidemics. As illustrated in \textbf{Figure~\ref{fig:biomedical_applications}e}, foundation models enable population-level applications such as real-time surveillance and epidemic forecasting by integrating multimodal health signals.

Foundation models have introduced new strategies for understanding large-scale public health trends. These models can process noisy, multimodal data streams and uncover relationships across social behavior, environmental patterns, and epidemiological signals. Their pretraining on heterogeneous datasets allows for generalization across population groups and regions.

\paragraph{Public Health Surveillance}
Public health surveillance involves monitoring disease trends and behavioral patterns using signals from clinical, social, and digital sources. Traditional surveillance systems often suffer from delays and sparse coverage.

Foundation models support improved sensitivity and responsiveness by analyzing diverse inputs, including social media, biomedical literature, and structured reports. Their ability to process unstructured data helps in identifying emerging threats and misinformation. PsychBERT~\cite{vajre2021psychbert} combines mental health literature with social content for mental health detection. Other models focus on detecting misinformation or monitoring virus transmission trends using large-scale social data. Representative public health foundation models for surveillance are summarized in Box~\ref{box:public_health}.

\paragraph{Epidemic Forecasting}
Epidemic forecasting involves predicting case trends, onset timing, peak severity, and epidemiological parameters such as reproduction numbers or attack rates. Conventional models require labor-intensive data preparation and are often limited in adaptability.

Foundation models address this by modeling temporal dynamics in health signals. These models can learn epidemic patterns from time series data and generalize across geographies and diseases. Temporal foundation models \cite{kamarthi2024large,gao2024units} are introduced for influenza-like illness forecasting; these models incorporate spatio-temporal dependencies and auxiliary variables to improve outbreak prediction. This shift enables more scalable and responsive epidemic modeling. Representative public health foundation models for surveillance and forecasting are summarized in Box~\ref{box:public_health}.

\section{Conclusion}
This survey focuses on the applications of foundation models in the health sciences. We reviewed the applications of foundation models in five areas: computational biology, drug development, clinical informatics, medical imaging, and public health. We hope that this survey will provide researchers and practitioners with a useful and detailed overview of foundation models in the health sciences, provide a convenient reference for relevant experts, and encourage future progress.

Looking ahead, the integration of Foundation Models within the health sciences promises to refine and accelerate existing processes and to pioneer new research and treatment methodologies. The journey toward fully realizing the potential of these models is intertwined with the continuous development of AI technologies, alongside the fostering of interdisciplinary collaborations among scientists, clinicians, and policymakers. As we navigate these challenges, the goal remains clear: to take advantage of the power of AI to improve health outcomes and pave the way for a new era of precision medicine and public health initiatives. The advancements in Foundation Models are not an end but a beginning, marking a pivotal moment in the evolving narrative of health science and artificial intelligence.





\citestyle{nature}
\bibliographystyle{unsrt} 
\bibliography{ref}

\clearpage

\onecolumn 
\setlength{\LTcapwidth}{\textwidth}

\appendix
\section*{Appendix}

\begin{nbeBox}{Foundation models for DNA languages in genome.}\label{box:genome}
\small
\begin{RowLinesLT}{p{0.14\textwidth} p{0.19\textwidth} p{0.12\textwidth} p{0.11\textwidth} p{0.11\textwidth} p{0.20\textwidth}}
\toprule
Work & Task & Architecture & Input & Output & Note \\
\midrule
BigBird\cite{zaheer2020big} (2020) & Question answering (QA), document summarization, promoter region prediction \& chromatin-profile prediction.  & Transformer& Long sequences (e.g., language, DNA) & Token-wise embeddings &
NA
\\\\
DNA-BERT\cite{ji2021dnabert} (2021)& Prediction of promoters, splice sites and transcription factor binding sites &Transformer & DNA sequence & Token-wise embedding & NA
\\\\          
GeneBERT\cite{mo2021multi} (2021) & Promoter classification, transaction factor binding sites prediction, disease risk estimation, splicing sites prediction & Transformer & Genome sequence \& 2D interaction matrix & Gene representation & Utilizes a 1D genome sequence and a 2D matrix representing interactions between transcription factors and genomic regions. 
\\\\ 
LOGO\cite{yang2022integrating} (2022)& Promoter identification, enhancer-promoter interaction prediction & Transformer& DNA sequence & token-wise embeddings & NA
\\\\         
LookingGlass\cite{hoarfrost2022deep} (2022)&  Identify novel oxidoreductase; predict enzyme optimal temperature; recognize reading frames of DNA sequence fragments & LSTM & DNA sequence & Token-wise embeddings & NA
\\\\
VIBE\cite{gwak2022vibe} (2022) & Eukaryotic viruses detection and classification & Transformer& Metagenome sequencing data & Token-wise embedding & A hierarchical BERT model to identify eukaryotic viruses using metagenome sequencing data and classify them at the order level.
\\\\         
INHERIT\cite{bai2022identification} (2022)& Phage identification  & Transformer& Bacteriophage genome sequences & Genome representation & NA
\\\\
Genomic Pre-trained Network (GPN)\cite{benegas2022dna} (2022)& Genome-wide variant effect predictions & CNN & Genomic sequence & Genome representation & NA
\\\\
DeepConsensus\cite{baid2023deepconsensus} (2023)& DNA sequence correction & Transformer& DNA sequence & Token-wise embeddings & DeepConsensus uses an alignment-based loss to train a gap-aware transformer–encoder for sequence correction.
 \\\\
 Nucleotide Transformer\cite{dalla2024nucleotide} (2024)& Molecular phenotype prediction & Transformer& Nucleotide sequence & Nucleotide representation & NA
\\\\ 
HyenaDNA\cite{nguyen2024hyenadna} (2024)& Chromatin profile prediction, species classification, regulatory elements identification & Hyena& DNA sequence & Token-wise embedding & Uses Hyena (sub-quadratic) to replace quadratic attention in transformers with implicit convolutions, enabling efficient scaling (up to 500x speedup) to 1M tokens with single-nucleotide-level resolution.  \\\\ 
GROVER (Genome Rules Obtained Via Extracted Representations)\cite{sanabria2024dna} (2024)& Genome element identification \& protein–DNA binding & Transformer & DNA sequence & Token-wise embedding & Trained on DNA sequences using byte-pair encoding. GROVER defines a vocabulary of tokens through a custom next-$k$-mer prediction task. \\\\ 
DNABERT-2\cite{zhou2023dnabert} (2024)& multi-species genome classification & Transformer& DNA sequence & Token-wise embeddings & NA
\\\\
Borzoi\cite{linder2023predicting} (2023) & DNA language model & Enformer (convolution + rransformer) & DNA sequence & DNA representation & Identifies key cis-regulatory patterns governing RNA expression and post-transcriptional regulation across normal tissues through attribution methods. 
\\\\ 
scooby\cite{hingerl2024scooby} (2024) & DNA language model & & DNA sequence & DNA representation &  NA
\\\\
Evo\cite{nguyen2024sequence} (2024) & Prediction \& design tasks from molecular to genome-scale & Hyena & DNA, RNA, protein & (DNA, RNA, protein) representation or sequence  &  The first examples of protein-RNA and protein-DNA co-design. 
\\\\
HiCFoundation\cite{wang2024generalizable} (2024)& Genome activity prediction & Transformer& 3D and 1D genome data & Genome representation & A Hi-C-based foundation model for integrative analysis of genome 3D architecture and its regulatory mechanisms. The first model that infers genome activity from the coarse genomic contact maps provided by Hi-C.\\
\bottomrule[1pt]
\end{RowLinesLT}
\end{nbeBox}

\clearpage 

\begin{nbeBox}{Foundation models for gene expression profiles by RNA.}\label{box:rna}
\small
\begin{RowLinesLT}{p{0.14\textwidth} p{0.1\textwidth} p{0.12\textwidth} p{0.11\textwidth} p{0.11\textwidth} p{0.30\textwidth}}
\toprule
Work & Task & Architecture & Input & Output & Note \\
\midrule 
scBERT\cite{yang2022scbert} (2022)  & RNA language model & Transformer & scRNA sequence & scRNA-seq representation & Pre-train BERT on massive unlabeled scRNA-seq data and fine-tuned on cell type annotation task.
\\\\ 
scFormer\cite{cui2022scformer} (2022) & RNA language model & Transformer & scRNA sequence & scRNA-seq representation & NA
\\\\
tGPT\cite{shen2022generative} (2022) & RNA language model & Transformer & scRNA sequence & scRNA-seq representation &  NA\\\\
scFoundation\cite{hao2023large} (2023) & RNA language model & Transformer & scRNA sequence & scRNA seq representation & NA 
\\\\ 
Geneformer\cite{theodoris2023transfer} (2023) & RNA language model & Transformer & scRNA sequence & scRNA seq representation & Pre-trained on 30M single-cell transcriptomes.
\\\\ 
scGPT\cite{cui2023scgpt} (2023) & RNA language model & Transformer & scRNA sequence & scRNA-seq representation &  Enhances scFormer\cite{cui2022scformer} with generative training techniques.
\\\\ 
sc-Long\cite{bai2024sclong} (2024) & RNA language model & Transformer & scRNA sequence & scRNA-seq representation & NA
\\\\
GenePT\cite{chen2024genept} (2024) & RNA language model & Transformer & Gene description & Gene embedding & NA
\\\\
SCSimilarity\cite{heimberg2024cell} (2024) & RNA language model & MLP & scRNA-seq & cell representation & Designed for rapid queries on similar cell profiles.\\\\
Cancer-Foundation\cite{theus2024cancerfoundation} (2024) & RNA language model & Transformer & scRNA sequence & scRNA-seq representation & Trained only on malignant cells, and its downstream task evaluates the generalizability to bulk RNA data.
\\ 
\bottomrule[1pt]
\end{RowLinesLT}
\end{nbeBox}

\clearpage

\begin{nbeBox}{Foundation models for protein structure prediction (a.k.a. protein folding) and protein design.}\label{box:protein}
\small
\begin{RowLinesLT}{p{0.14\textwidth} p{0.1\textwidth} p{0.12\textwidth} p{0.11\textwidth} p{0.11\textwidth} p{0.30\textwidth}}
\toprule
Work & Task & Architecture & Input & Output & Note \\
\midrule 
AlphaFold2\cite{jumper_2021_highly} (2021)  & Protein structure prediction & Transformer & Protein sequence & Protein structure & \textbf{Milestone} work in single-chain protein structure prediction, dominating CASP. 
\\\\
RoseTTAFold\cite{baek2021accurate} (2021) & Protein structure prediction & Transformer & Protein sequence & Protein structure & Protein monomer prediction, 3-track network architecture (1D sequence level, 2D distance map level and 3D coordinate level).
\\\\
AlphaFold-Multimer\cite{evans2021protein} (2021) & Protein structure prediction & Transformer & Protein sequence & Protein structure & Predict structures for protein multimers.
\\\\
ProtTrans\cite{elnaggar2021prottrans} (2021) & Protein language model& Transformer& Protein sequence& Protein representation & NA
\\\\
ProBERT\cite{brandes2022proteinbert} (2022) & Protein representation learning& Transformer&Protein sequence& Protein representation& NA
\\\\ 
ESM\cite{lin2022language} (2022)  & Protein language model& Transformer&Protein sequence& Protein representation & The Evolutionary Scale Modeling (ESM) family of protein language models, including ESM-1v\cite{meier2021language}, ESM-1b\cite{rives2019biological}, and ESM-MSA\cite{rao2021msa} and etc. 
\\\\
ESM-IF1\cite{hsu2022learning} (2022) & Protein design & GNN, Transformer & Protein structure& Protein sequence & Backbone structure to sequence design conditioned on sequences.
\\\\
OmegaFold\cite{wu2022high} (2022) & Protein structure prediction& Transformer & Protein sequence & Protein structure & Uses protein language modeling to replace MSA, less accurate but faster. 
\\\\
RFDiffusion\cite{watson2023novo} (2023) & Protein design & Diffusion model& Protein structure& Protein structure & A generative diffusion model from structure to structure for protein design. 
\\\\
EvoDiff\cite{alamdari2023protein} (2023) & Protein design & Diffusion model & Protein sequence& Protein sequence& Controllable sequence-level protein design diffusion model. 
\\\\
Chroma\cite{ingraham2023illuminating} (2023) & Protein design & GNN, diffusion &Programmable protein condition& Protein sequence, structure & Achieving programmable generation with user-specified properties. 
\\\\ 
ProtHyena\cite{zhang2024prothyena} (2024) & Protein language model & Hyena & Protein sequence& Protein representation & Adopts Hyena operator for efficient scaling. 
\\\\
xTrimoPGLM\cite{chen2024xtrimopglm} (2024) &  18 protein understanding tasks & Transformer & Protein sequence& Protein representation & NA
\\\\ 
ESM3\cite{hayes2024simulating} (2024) & Protein language model& Transformer& Protein sequence, structure, function& Protein sequence, structure, function & NA 
\\\\
Protein Generator\cite{lisanza2024multistate} (2024) & Protein design & Transformer, diffusion & Protein sequence & Protein sequence, structure & A RoseTTAFold-based sequence diffusion model that simultaneously generates protein sequences and structures.
\\\\
RoseTTAFold All-Atom\cite{krishna2024generalized} (2024) & All-atom structure prediction & Transformer& All-atom sequences, ligands, bonds & All-atom structure & Generalized model for all-atom prediction including protein, nucleic acid, and other small molecules.
\\\\ 
AlphaFold3\cite{abramson2024accurate} (2024) & All-atom structure prediction & Transformer \& diffusion & All-atom sequences, ligands, bonds & All-atom structure & State-of-the-art all-atom prediction method.
\\\\ 
Boltz-1\cite{wohlwend2024boltz} (2024) & All-atom structure prediction & Transformer \& diffusion & All-atom sequences, ligands, bonds & All-atom structure & AlphaFold3-level accuracy, open-source. 
\\\\
Chai-1\cite{boitreaud2024chai} (2024) & All-atom structure prediction & Transformer \& diffusion & All-atom sequences, ligands, bonds & All-atom structure & Can also predict structures with sequence only. \\
\bottomrule[1pt] 
\end{RowLinesLT}
\end{nbeBox}

\clearpage

\begin{nbeBox}{Foundation models for drug molecular representation learning.}\label{box:molecular}
\small
\begin{RowLinesLT}{p{0.12\textwidth} p{0.16\textwidth} p{0.1\textwidth} p{0.11\textwidth} p{0.11\textwidth} p{0.30\textwidth}}
\toprule
Work & Task & Architecture & Input & Output & Note \\
\midrule
Wang et al.\cite{wang2021chemical}  (2021)& Molecular representation learning & GNN & Molecule, chemical reaction & Molecular embedding & Integrates chemical reaction constraints to enhance molecular embeddings: forcing the sum of reactant embeddings equals the sum of product embeddings. 
\\\\ 
Su et al.\cite{su2022molecular} (2022) & Graph-text/text-graph retrieval, molecule captioning, property prediction, text-based drug design & Transformer \& GNN & Molecular graph, molecular diagram, text & Molecular embedding & NA
\\\\
MolT5\cite{edwards2022translation} (2022) & Molecule captioning \& text-based drug design & Transformer & Molecule or text & Text or molecule & NA
\\\\ 
Zeng et al.\cite{zeng2022deep} (2022) & Property prediction \& biomedical relation extraction& Transformer& Text, molecular structure & Text, molecule & Integrates molecule and text through unsupervised meta-knowledge learning. 
\\\\
MolKD\cite{zeng2023molkd} (2023) & Property prediction&Transformer & Chemical reaction, reaction yield & Molecule representation & MolKD distilled knowledge from a teacher model trained on reaction data to a student model. Also, MolKD integrates reaction yield information during pre-training to measure reactant-product transformation efficiency.  
\\\\
CLAMP\cite{seidl2023enhancing} (2023) & Property prediction & Transformer & Text \& molecule & Text description & NA
\\\\
MolFM\cite{luo2023molfm} (2023) & Property prediction & Transformer & Molecular graphs & Molecular representation & Two-step pretraining: (1) self-supervised learning for chemical structure representation; (2) multi-task learning for biological information integration.
\\\\ 
MoleculeSTM\cite{liu2023multi} (2023) & Drug design, biological property prediction, instruction adaptation & GNN (molecule), transformer (molecule, text) & Molecule \& text & Molecule, text &NA. 
\\\\
InstructMol\cite{cao2023instructmol} (2023) & Property prediction& Multimodal LLM & Molecule \& text & Molecule or text & NA
\\\\
BioT5\cite{pei2023biot5} (2023) & Molecule \& protein property prediction, drug-target / protein-protein interaction, molecule captioning, text-based drug design & Transformer & SELFIES, protein, text & Text, or molecule & Uses SELFIES strings for 100\% molecule validity; extracts contextual knowledge from unstructured biological literature.
\\\\
BioT5+\cite{pei2024biot5+} (2024) & Molecule-to-text, text-to-molecule & Transformer & Text, molecule & Text or molecule & Enhance BioT5 by integrating text and molecular representations. 
\\\\ 
MV-Mol (multi-view molecule) \cite{luo2024learning} (2024) & Molecular representation learning & Multimodal fusion architecture &  Molecular structure, text, knowledge graph & Molecular embedding & NA 
\\\\ 
UniMoT (Unified Molecule-Text LLM)\cite{zhang2024unimot} (2024) & Molecule-to-text \& text-to-molecule generation & Transformer & Molecule or text & Text or molecule & Uses a Vector Quantization-driven tokenizer and a Q-Former to bridge molecule and text modalities. \\
\bottomrule[1pt]
\end{RowLinesLT}
\end{nbeBox}

\clearpage

\begin{nbeBox}{Foundation model for pharmaceutical property prediction.}\label{box:PPP}
\small
\begin{RowLinesLT}{p{0.14\textwidth} p{0.1\textwidth} p{0.12\textwidth} p{0.11\textwidth} p{0.11\textwidth} p{0.30\textwidth}}
\toprule
Work & Task & Architecture & Input & Output & Note \\
\midrule 
SMILES-BERT\cite{wang2019smiles} (2019) & Property prediction & Transformer & SMILES strings & Token-wise embedding & NA
\\\\
MolE\cite{mendez2022mole} (2022) & Property prediction & Transformer & Molecular graphs & Molecular representation & DeBERTa architecture. two-step pretraining: self-supervised learning for chemical structure representation and multi-task learning for biological information integration. 
\\\\
ChemBERTa-2\cite{ahmad2022chemberta} (2022) & Property prediction & Transformer & SMILES strings & Token-wise embedding & Pretrain on 77M unlabeled SMILES strings from PubChem, one of the largest molecular pretraining datasets to date. 
\\\\
BAITSAO\cite{zhao2024building} (2024) & Drug synergy prediction& Transformer& Drug combination, cell lines & Drug synergy & NA
\\\\
ActFound\cite{feng2024bioactivity} (2024) & Bioactivity prediction & Transformer & Molecular structure & Molecular representation & Trained on 1.6M experimentally measured bioactivities from 35K in ChEMBL, ActFound designs pairwise learning and meta-learning to capture relative bioactivity differences between compounds within the same assay, overcoming cross-assay incompatibility. 
\\\\
SMILES-Mamba\cite{xu2024smiles} (2024) & ADMET property prediction & Mamba & SMILES string & Molecular representation & NA 
\\\\
ChemFM\cite{cai2024foundation} (2024) & Property prediction & Transformer & Molecular structure & Molecular representations & Up to 3B parameters, pre-trained on 178M molecules using self-supervised causal language modeling to generate generalizable molecular representations. Supports full-parameter and parameter-efficient fine-tuning.  
\\\\
Graph Transformer Foundation Model (GTFM)\cite{mizera2024graph} (2024) & Property prediction & Graph transformer & Molecular graph & Molecular representations & NA  \\
\bottomrule[1pt]
\end{RowLinesLT}
\end{nbeBox}

\clearpage

\begin{nbeBox}{Foundation models for drug repurposing (a.k.a., drug reuse, drug repositioning).}\label{box:repurposing}
\small
\begin{RowLinesLT}{p{0.12\textwidth} p{0.12\textwidth} p{0.12\textwidth} p{0.11\textwidth} p{0.11\textwidth} p{0.30\textwidth}}
\toprule
Work & Task & Architecture & Input & Output & Note \\
\midrule 
Zhu et al.~\cite{zhu2020knowledge} (2020) & Drug Repurposing & GNN &Drug \& disease & Drug-disease association score & NA 
\\\\
KG-Predict~\cite{gao2022kg} (2022) & Drug Repurposing & GNN &Drug \& disease & Drug-disease association score & Combines GCN and InteractE that processes embeddings using 3D tensor convolution to capture heterogeneous interactions.
\\\\
DREAMwalk~\cite{bang2023biomedical} (2023) & Drug Repurposing & GNN & Drug, gene, \& disease & Drug-disease association score & NA 
\\\\
TxGNN~\cite{huang2024foundation} (2024) & Drug Repurposing & GNN & Diseases, drugs, proteins, \& pathways & Drug-disease association score\& explanation& Designs an Explainer module via multi-hop paths in the knowledge graph for interpretability.
\\\\
HGTDR~\cite{gharizadeh2024hgtdr} (2024)& Drug Repurposing & Graph transformer &Drug \& disease & Drug-disease association score & NA \\
\bottomrule[1pt]
\end{RowLinesLT}
\end{nbeBox}

\clearpage

\begin{nbeBox}{Foundation models for clinical trial prediction.}\label{box:trial}
\small
\begin{RowLinesLT}{p{0.12\textwidth} p{0.12\textwidth} p{0.12\textwidth} p{0.11\textwidth} p{0.11\textwidth} p{0.30\textwidth}}
\toprule
Work & Task & Architecture & Input & Output & Note \\
\midrule
HINT\cite{fu2022hint} (2022) & Clinical trial outcome prediction & GNN & Drug, disease code, text & Trial outcome & NA
\\\\
inClinico\cite{aliper2023prediction} (2023) & Trial outcome prediction &Transformer & Multiomics data, trial design, drug properties & Trial outcome &NA
\\\\
HINT-UQ\cite{chen2024uncertainty} (2024) & Trial outcome prediction & GNN & Drug, disease code, trial protocol (text) & Trial outcome & HINT-UQ quantifies uncertainty for reliable prediction using selective classification. 
\\\\
TrialDura\cite{yue2024trialdura} (2024) & Trial duration prediction & Transformer & Disease names, drug molecules, trial phases, \& eligibility criteria & Trial duration & NA
\\\\
LIFTED\cite{zheng2024multimodal} (2024) & Trial outcome prediction & Sparse mixture-of-expert & Drug, disease, trial protocol & Trial outcome & LIFTED uses a sparse Mixture-of-Experts framework to identify cross-modal patterns and provide explanations using a shared expert model and dynamic weighting mechanism. 
\\\\
CTP-LLM\cite{reinisch2024ctp}  (2024) & Trial phase transition prediction & GPT & Trial design document & Trial phase transition & NA
\\\\
TrialEnroll\cite{yue2024trialenroll} (2024) & Trial enrollment prediction & Deep cross network & Eligibility criteria & Trial enrollment status & NA
\\\\
ClinicalAgent\cite{yue2024clinicalagent} (2024) & Trial outcome prediction & GPT4 & Drug, disease and text & Trial outcome & GPT4-based multi-agent system that integrates LEAST-TO-MOST and ReAct reasoning. \\ 
\bottomrule[1pt]
\end{RowLinesLT}
\end{nbeBox}

\clearpage
    
\begin{nbeBox}{Foundation models for clinical summarization.}\label{box:summarization}
\small
\begin{RowLinesLT}{p{0.14\textwidth} p{0.1\textwidth} p{0.12\textwidth} p{0.09\textwidth} p{0.11\textwidth} p{0.30\textwidth}}
\toprule
Work & Task & Architecture & Input & Output & Note \\
\midrule 
Sotudeh et al.\cite{sotudeh2020attend} (2020) & Clinical summarization & LSTM & Clinical document & Summary & NA
\\\\
BioBERTSum\cite{du2020biomedical} (2020) & Clinical summarization &  Transformer & Clinical document & Summary & Pretrained BERT as encoder, followed by finetuning.  
\\\\
KeBioSum\cite{xie2022pre} (2022) & Clinical summarization & Transformer &Text  & Summary and sentence classification result& NA
\\\\
COVIDSum \cite{cai2022covidsum} (2022) & Clinical summarization &Transformer \& Graph attention network&  Medical articles & Summary & NA \\\\
Radiology-LLaMA2\cite{liu2023radiology} (2023) &Clinical summarization &	Transformer	& Radiology report & Summary & NA\\\\    
MRC-Sum\cite{li2023mrc} (2023) & Clinical summarization & Transformer & Text & Summary and extracted information &NA \\ 
\bottomrule[1pt]
\end{RowLinesLT}
\end{nbeBox}

\clearpage
    
\begin{nbeBox}{Foundation models for clinical QA.}\label{box:clinicalqa}
\small
\begin{RowLinesLT}{p{0.14\textwidth} p{0.1\textwidth} p{0.12\textwidth} p{0.09\textwidth} p{0.11\textwidth} p{0.30\textwidth}}
\toprule
Work & Task & Architecture & Input & Output & Note \\
\midrule 
Yoon et al.\cite{yoon2019pre} &Clinical QA& Transformer & Text &Extracted terms and entities, classification result  &NA
\\\\ 
BioMedBERT\cite{chakraborty2020biomedbert} (2020) & Clinical QA, named entity recognition (NER) & Transformer & Text \& entity pair& QA, NER result, relation extraction result  & NA
\\\\
Rawat et al.\cite{rawat2020entity} (2020) &Clinical QA & Transformer & Text & QA, structured semantic representation & NA
\\\\
Yan et al. \cite{yan2022remedi}(2022) (2022) & Clinical QA & Transformer & Text \& 	medical knowledge base  &Intent-slot-value triplets, action-slot-value pairs, QA & NA
\\\\
DAPO\cite{li2023dialogue} (2023)& Clinical QA&Transformer &Text &Prediction scores, ranked responses & DAPO considers dialogue-specific features such as coherence, specificity, and informativeness.
\\\\
ClinicalGPT\cite{wang2023clinicalgpt} (2023) &Clinical QA &	GPT & Text &  Medical diagnoses, treatment recommendations, summary	&  NA
\\\\ 
Deid-GPT\cite{liu2023deid} (2023) &Clinical QA&	GPT & Text &De-Identified medical text & The model replaces identifiable data following HIPAA guidelines.
\\\\
ChiMed-GPT\cite{tian2023chimed} (2023) & Information extraction, QA, dialogue generation & GPT & Text & Text & Specifically trained for Chinese medicine.
\\\\
Med-PaLM\cite{singhal2023large} (2023)&Clinical QA & Transformer & Text, image & text, image & Multimodal biomedical AI model that can answer complex questions, generate reports, and classify images, developed by Google. 
\\\\
Med-PaLM 2\cite{singhal2023towards} (2023)&Clinical QA &Transformer &Text & QA and adversarial evaluation result& The first model to pass the US Medical Licensing Examination. 
\\\\    
MedAgents\cite{tang2023medagents} (2023)&Clinical QA &Transformer  & Image \& text  & QA & LLM agent, training-free, accessing external knowledge. \\ 
Pmc-LLaMA\cite{wu2024pmc} (2024)& Clinical QA&Transformer &Text & QA, summary, relation extraction, classification result, and diagnosis& NA
\\\\
Me-LLaMA\cite{xie2024me} (2024)&  Clinical QA & Transformer  & Text &QA, summary, classification result &NA \\\\
\bottomrule[1pt]
\end{RowLinesLT}
\end{nbeBox}

\clearpage

\begin{nbeBox}{Foundation models for patient representation.}\label{box:patient_representation}
\small
\begin{RowLinesLT}{p{0.14\textwidth} p{0.1\textwidth} p{0.12\textwidth} p{0.11\textwidth} p{0.11\textwidth} p{0.30\textwidth}}
\toprule
Work & Task & Architecture & Input & Output & Note \\
\midrule 
BEHRT\cite{li2020behrt} (2020) & EHR concept \& visit representation & Transformer & EHR & Clinical prediction and patient representation & NA
\\\\
CEHR-BERT\cite{pang2021cehr} (2021) & EHR concept \& visit representation & Transformer & EHR & EHR representation & Uses artificial time tokens and Fourier transform-based time2vec encoding to represent seasonal or age-related patterns.
\\\\
Hi-BEHRT\cite{li2022hi} & EHR concept \& visit representation& Transformer & EHR & EHR representation & Hierarchical BERT: the local feature extractor captures short-term dependencies using sliding window segmentation; the global feature aggregator learns long-term dependencies over EHR histories.
\\\\
ExBEHRT\cite{rupp2023exbehrt} (2023) &EHR concept \& visit representation& Transformer & EHR & Clinical prediction, patient representation and patient cluster & NA
\\\\  
M-BEHRT\cite{mbaye2024multimodal} (2024) & EHR concept \& visit representation & Transformer & EHR & Breast cancer prognosis prediction and patient representations & NA
\\\\
GT-BEHRT\cite{poulain2024graph} (2024) &EHR concept \& visit representation & Graph transformer & EHR & Patient representation and clinical risk prediction & NA
\\\\
Claimsformer\cite{gerrard2024claimsformer} (2024) &EHR concept \& visit representation& Transformer & EHR & Chronic condition prediction and patient representations & NA
\\\\
HEART\cite{huang2024heart} (2024) &EHR concept \& visit representation& transformer & EHR & Clinical risk prediction and patient representation & Applies type-specific transformations to medical entities to learn relationship-specific attention biases to prioritize clinically relevant interactions.
\\\\ 
HERBERT\cite{moore2024healthrecordbert} (2024) & Risk stratification in chronic kidney disease & Transformer & EHR & Disease risk stratification and patient representation &  NA\\\\
EHRMamba\cite{fallahpour2024ehrmamba} (2024)  & EHR concept \& visit representation & Mamba & EHR & EHR forecasting, clinical risk prediction, patient representation & EHRMamba uses state-space models instead of transformers for linear-time sequence modelling. EHRMamba can
   process sequences up to 3x longer.
\\\\
TAME\cite{zeng2021transformer} (2021) & Patient Representation&Transformer & EHR & Patient representation, patient subtyping and clinical risk prediction  & NA\\\\   
RAPT\cite{ren2021rapt} (2021) & Patient Representation & Transformer &EHR &Clinical prediction and clinical decision support & NA\\\\
Claim-PT\cite{zeng2022pretrained} (2022) & Patient representation & GPT  & EHR & & NA\\\\
Guo et al.\cite{guo2023ehr} (2023) & Patient Representation& Transformer \& GRU & EHR &Clinical risk prediction and patient representation & NA
\\\\
Foresight\cite{hofmann2024foresight} (2024) & Patient Representation &GPT  &EHR&Biomedical events forecast, risk stratification and virtual trial result & NA
\\\\
CEHR-GAN-BERT\cite{poulain2022few} (2022) & Predictive phenotyping & Transformer \& GAN & EHR & Patient representation and predicted clinical outcomes & The generator mimics BERT-derived EHR representations; the discriminator distinguishes the generated one from the real one. 
\\\\
Hur et al.\cite{hur2022unifying} (2022) & Predictive phenotyping& Transformer & EHR & Patient representation and predicted clinical outcomes&  NA
\\\\    
PSN\cite{navaz2022novel} (2022) &Patient Subphenotyping and Similarity Measurement &Transformer \& CNN \& LSTM  &EHR&Patient similarity score and disease risk & PSN uses similarity network fusion to integrate structured EHR data with unstructured clinical narratives.
\\\\
DAPSNet\cite{wu2023dual} (2023) &Patient Subphenotyping and Similarity Measurement&Transformer &EHR  & Drug recommendation & NA
\\\\
ExBEHRT\cite{rupp2023exbehrt} (2023) & Patient Subphenotyping and Similarity Measurement & Transformer & EHR&Patient group, risk score and mortality risk prediction &  NA\\\\
TransTEE\cite{zhang2022exploring} (2022) & Treatment Effect Estimation & Transformer & EHR & Estimated treatment effects & NA
\\\\
Cure\cite{liu2022cure} (2022) & Treatment Effect Estimation & Transformer & Patient data& Estimated treatment effects & Encodes structured observational patient data and incorporates covariate type and time into patient embeddings from unlabeled large-scale datasets.\\ \\ 
\bottomrule[1pt]
\end{RowLinesLT}
\end{nbeBox}

\clearpage

\begin{nbeBox}{Foundation models for pathology.}\label{box:pathology}
\small
\begin{RowLinesLT}{p{0.14\textwidth} p{0.1\textwidth} p{0.12\textwidth} p{0.09\textwidth} p{0.11\textwidth} p{0.30\textwidth}}
\toprule
Work & Task & Architecture & Input & Output & Note \\
\midrule 
MI-Zero\cite{lu2023visual} (2023) & Pathology & & Image \& text &Cancer subtyping and region-of-interest identification results  &  NA\\\\
PLIP\cite{huang2023visual} (2023)& Pathology &	Transformer	 &Image \& text & Feature representation, classification result& NA\\\\
CITE\cite{zhang2023text}(2023) & Pathology &	Transformer &Image \& text &Feature representation, classification result & NA\\\\
Virchow\cite{vorontsov2023virchow}(2023) & Pathology & Transformer & Image& Cancer detection result& NA \\\\
UNI\cite{chen2024towards}(2023) & Pathology &	Transformer  &Image & Segmentation mask, Cancer detection, grading, and subtyping results & NA\\\\
PathChat\cite{lu2023foundational}(2023)& Pathology &	Transformer  &Image \& text & Prediction results, report, QA &NA \\\\
CHIEF\cite{wang2024pathology}(2024)& Pathology & Transformer  & Image \& text& Detection, classification, and prediction result & NA
\\\\
RudolfV\cite{dippel2024rudolfv} (2024)& Pathology & Transformer& Image \& text&Feature representation, cell segmentation mask, biomarker scoring, and rare disease case retrieval result &NA
\\\\
PANTHER\cite{song2024morphological}(2024) & Pathology	& Transformer & Image& Feature representation, cancer subtyping, survival outcome prediction&PANTHER reduces pathology whole-slide image patches into a compact set of morphological prototypes for efficient slide representation.\\\\
Jaume et al.\cite{jaume2025multistain}(2024) & Pathology &	Transformer  & Image \& text& Feature representation, molecular subtyping, prognostic prediction result & NA \\\\
XLIP\cite{wu2024xlip} (2024)&	Pathology&	Transformer  &Image \& text & & XLIP's Attention-Masked Image Modelling module masks image features that are highly responsive to textual features. The Entity-Driven Masked Language Modelling module enhances medical-specific features. \\\\
Prov-GigaPath \cite{xu2024whole} (2024) & Pathology & Transformer & Image & Feature embedding, cancer subtyping, pathomics classification & Prov-GigaPath uses a DINOv2  as a tile encoder to extract local feature and a LongNet to process tens of thousands of image tiles per slide for ultra-large-context modeling. \\\\
MUSK \cite{xiang2025vision} (2025) & Pathology &Transformer& Image \& clinical report & Feature representation, QA, molecular biomarker prediction, cancer prediction, immunotherapy response prediction &NA \\
\bottomrule[1pt]
\end{RowLinesLT}
\end{nbeBox}

\clearpage 
    
\begin{nbeBox}{Foundation models for radiology.}\label{box:Radiology}
\small
\begin{RowLinesLT}{p{0.14\textwidth} p{0.1\textwidth} p{0.12\textwidth} p{0.09\textwidth} p{0.11\textwidth} p{0.30\textwidth}}
\toprule
Work & Task & Architecture & Input & Output & Note \\
\midrule 
CheXzero\cite{tiu2022expert} (2022) & Radiology &	Transformer  &Image \& text & classification result and auxiliary prediction result&	NA \\\\
RadFM\cite{wu2023towards} (2023)	&Radiology&	Transformer & Image \& text &Feature representation, classification result,report,  QA &NA\\\\
MedBLIP\cite{chen2024medblip} (2024) &	Radiology&	Transformer  & Image \& text  & Feature representation, classification result, QA & 	MedBLIP uses a MedQFormer module to bridge the gap between 3D medical images to the pre-trained model.\\\\
BioVil-T\cite{bannur2023learning} (2023) &Radiology&	Transformer  & Image \& text & Feature representation, classification result, temporal sentence similarity  & NA\\\\
PTUnifier\cite{chenptunifier} (2023) &Radiology&	Transformer  &Image \& text & Feature representation, classification result, summarization, question answer & NA \\\\
KoBO\cite{chen2023knowledge} (2023) &Radiology&	Transformer \& CNN \& GNN	 & Image \& text & Feature representation, classification result, segmentation mask, and semantic analysis. & The KoBo framework integrates clinical knowledge to improve semantic consistency and introduces an unbiased, open-set knowledge representation to handle noisy samples. \\\\
ELIXR\cite{xu2023elixr} (2023) & Radiology&	 & & 	& \\\\
MaCO\cite{huang2024enhancing} (2024) &Radiology&	Transformer	& Image \& text & Classification, segmentation, detection results &  Maco incorporates a correlation weighting mechanism to refine the alignment between masked X-ray image patches and their associated reports. \\\\
Clinical-BERT\cite{yan2022clinical}&Radiology&	Transformer \& CNN	& Image \& text & Report  &Clinical-BERT employs Masked Medical Subject Headings (MeSH) Modeling where MeSH is a semantic component in radiograph reports, and Image-MeSH Matching to align visual features with MeSH terms using a two-level sparse attention mechanism.\\\\
SAMed\cite{zhang2023customized} (2023)&Radiology&	Transformer & Image \&text & Segmentation mask	& SAMed utilizes a low-rank adaptation strategy, updating the SAM image encoder, prompt encoder, and mask decoder using labelled medical datasets.\\\\
Xraygpt\cite{thawkar2023xraygpt}&Radiology&	Transformer 	& Image \& text & Report and question answer& NA	\\\\
Chatcad\cite{wang2023chatcad} (2023) &Radiology&	Transform  	&Image \& text	& Report and advice& NA \\\\
3D-CT-GPT\cite{chen20243d} (2024) & Radiology & Transformer & Image \& text& Report &NA\\\\
ZePT\cite{jiang2024zept} (2024) &Radiology&	Transformer  & Image \& text& Segmentation mask and anomaly map& ZePT uses a two-stage training approach: first, learning fundamental queries for organ segmentation via object-aware feature grouping to capture organ-level features, and second, refining advanced queries with auto-generated visual prompts for detecting unseen tumours.\\\\
miniGPT-Med\cite{alkhaldi2024minigpt}&Radiology&	Transformer	& Image \& text & Report, bounding box and QA &NA\\\\
MAIRA-2\cite{bannur2024maira} (2024)&Radiology&	Transformer	& Image \& text & Grounded and non-grounded report&NA\\\\
BrainSegFounder\cite{cox2024brainsegfounder} (2024) &Radiology&	Transformer \& U-Net& Image& Segmentation mask &NA \\\\
ChEX\cite{muller2025chex} &Radiology	& Transformer 	&Image \& text & Bounding box \& description &Chest X-Ray Explainer (ChEX) integrates textual prompts and bounding boxes to allow the interpretation of specific anatomical regions and pathologies. \\
\bottomrule[1pt]
\end{RowLinesLT}
\end{nbeBox}

\clearpage 

\begin{nbeBox}{Foundation models for retinal image.}\label{box:retinal_image}
\small
\begin{RowLinesLT}{p{0.14\textwidth} p{0.1\textwidth} p{0.12\textwidth} p{0.09\textwidth} p{0.11\textwidth} p{0.30\textwidth}}
\toprule
Work & Task & Architecture & Input & Output & Note \\
\midrule 
RETFound\cite{zhou2023foundation} (2023) &Retinal Images& Transformer \& CNN& Image \& text& Feature presentation, symptom classification result. & NA
\\\\ 
DeepDR Plus\cite{dai2024deep}&Retinal Images& CNN & Image \& metadata & Progression and risk score& DeepDR Plus is designed to predict the time to diabetic retinopathy progression over five years using only fundus images. 
\\\\
KeepFIT\cite{wu2024mm} (2024) &Retinal Images& Transformer \& CNN & Image \& text & Feature presentation, symptom classification result, and image captioning & The model integrates Fundus Image-Text expertise through image similarity-guided text revision and a mixed training strategy. 
\\\\
RetiZero\cite{wang2024common}  (2024) &Retinal Images& Transformer \& MAE &Image \& text&Feature presentation, symptom classification result, Image Retrivel & NA
\\\\ 
RET-CLIP\cite{du2024ret} (2024) &Retinal Images& Transformer  & Image \& text& Feature presentation, symptom classification results.& RET-CLIP uses a tripartite optimization strategy that considers both eyes, and patient-level data, aligning with real-world clinical scenarios.
\\\\
FLAIR\cite{silva2025foundation} (2025) &Retinal Images& Transformer \& CNN & Image & Symptom classification result& NA\\
\bottomrule[1pt]
\end{RowLinesLT}
\end{nbeBox}

\clearpage

\begin{nbeBox}{Foundation models for multimodal medical imaging.}\label{box:medical_imaging}
\small
\begin{RowLinesLT}{p{0.14\textwidth} p{0.11\textwidth} p{0.12\textwidth} p{0.09\textwidth} p{0.11\textwidth} p{0.30\textwidth}}
\toprule
Work & Task & Architecture & Input & Output & Note \\
\midrule 
BiomedGPT\cite{zhang2023biomedgpt}& Multi-modality & Transformer & Image \& text& Feature representation \& QA&NA
\\\\    
BiomedCLIP\cite{zhang2023biomedclip} &	Multi-modality&	Transformer	&Image \& text& Feature representation \& QA &NA
\\\\
Med-Flamingo\cite{moor2023med} & Multi-modality &Transformer&Image \& text& Report \& QA &NA
\\\\
MedSAM\cite{ma2024segment} & Multi-modality& Transformer &Image&Segmentation mask&	NA
\\\\
SAM-Med2D\cite{cheng2023sam}&Multi-modality&	Transformer &Image & Segmentation mask&  SAM-Med2D incorporates more diverse prompts: bounding boxes, points, and masks.
\\\\
LVM-Med\cite{mh2024lvm} & Multi-modality	&	Transformer \& GNN & Image &Feature representation, segmentation mask, detection and classification results & In LVM-Med, two sets of feature embeddings are produced using transformers to construct a graph neural network each representing nodes and edges with second-order graph matching algorithm.   \\\\
AutoSAM\cite{hu2023efficiently} & Segmentation &	Transformer & Image& Segmentation mask& NA
\\\\
Med-SA\cite{wu2023medical} & Segmentation	& Transformer	& Image &  &  Med-SA employs space-depth transpose to extend SAM’s 2D capabilities to 3D medical images and a Hyper-Prompting Adapter for prompt-conditioned adaptation. 
\\\\
Llava-med\cite{li2024llava}&	Multi-modality &Transformer & Image \& text& Report \& QA & LLaVA-Med first aligns with biomedical vocabulary using figure-caption pairs, then learns conversational semantics through instruction-following data, mimicking a layperson’s gradual acquisition of biomedical knowledge. \\\\ 
Visual Med-Alpaca\cite{shu2023visual} &	Multi-modality	& Transformer  &Image \& text & Classification result \& QA & NA\\\\
RELU\cite{xia2024rule}	& Multi-modality &	Transformer  & Image \& text & & The RULE framework includes a calibrated retrieval strategy to optimize factual risk and a fine-tuned preference dataset to improve retrieval-augmented generation.  \\\\
MA-SAM\cite{du2024ret}& Multi-modality&	Transformer  & Image& Segmentation mask & MA-SAM adapts SAM’s 2D backbone to handle volumetric and temporal information. It also integrates 3D adapters to extract 3D features while preserving pre-trained 2D weights through efficient fine-tuning. \\\\
UniMed-CLIP\cite{uzair2024unimed}& Multi-modality&	Transformer	 &Image \& text & Feature representation, classification result &NA  \\\\
HuatuoGPT-Vision\cite{chen2024huatuogpt}&	Multi-modality&	Transformer  &Image \& text  &  Feature representation, QA &NA \\\\
RadEdit\cite{perez2025radedit} & Multi-modality &	Diffusion  & Image \& text & Synthetic datasets & RadEdit uses generative image editing to simulate dataset shifts and diagnose failure modes of biomedical vision models. The model uses multiple image masks to constrain edits and ensure consistency. \\\\
BiomedParse \cite{zhao2024foundation} (2024)& Multi-modality & Transformer & Image \& text \& semantic label & Segmentation mask, object recognition and clinical detection result &NA\\
\bottomrule[1pt]
\end{RowLinesLT}
\end{nbeBox}

\clearpage

\begin{nbeBox}{Foundation models for public health surveillance through multimodal and heterogeneous data.}\label{box:public_health}
\small
\begin{RowLinesLT}{p{0.14\textwidth} p{0.13\textwidth} p{0.12\textwidth} p{0.11\textwidth} p{0.11\textwidth} p{0.30\textwidth}}
\toprule
Work & Task & Architecture & Input & Output & Note \\
\midrule 
Luo et al.\cite{luo2021deep} (2021) & Virus spreading information & Transformer &Clinical notes \& Social media posts& Symptoms& NA\\\\
COVID-19 Surveiller\cite{jiang2022covid} (2022) & Virus spreading information &  Transformer \& GNN& Social media posts & COVID-19 event prediction&NA\\\\
PsychBERT\cite{vajre2021psychbert} (2021) & Mental Health Surveillance &  Transformer &Social media posts & Mental condition detection & NA\\\\
PHS-BERT\cite{naseem2022benchmarking}  (2022)  &Mental Health Surveillance & Transformer & Social media posts & Mental condition detection & NA\\\\
Saha et al.\cite{saha2022social} (2022) & Mental Health Surveillance &Auto-Regressive Integrated Moving Average Model  & Social media posts & Mental condition detection & NA \\\\
MentaLLaMA\cite{yang2024mentallama} (2024) & Mental Health Surveillance & Transformer &Social media posts & Mental condition detection&   NA\\\\
Deka et al. \cite{deka2022improved} (2022)& Misinformation detection &Transformer &  Medical articles & Classification result & NA \\\\
Vec4Cred \cite{upadhyay2023vec4cred} (2023) & Misinformation detection &CNN \& LSTM \& Attention &  Web textural content & Classification result &  NA\\\\
Upadhyay et al. \cite{upadhyay2023leveraging} (2023) & Misinformation detection &CNN \& LSTM \& Attention &  Web textural content & Classification result & NA \\\\
\bottomrule[1pt]
\end{RowLinesLT}
\end{nbeBox}

\end{document}